\documentclass[journal]{IEEEtran}

\usepackage[linesnumbered, ruled,vlined]{algorithm2e}
\usepackage{times}
\usepackage{soul}
\usepackage{url}
\usepackage{graphicx}
\usepackage{amsmath}
\usepackage{booktabs}
\usepackage{nicematrix}
\usepackage{amsfonts}
\usepackage{array}
\usepackage{stfloats}
\usepackage{color}
\usepackage[tight,footnotesize]{subfigure}
\usepackage{adjustbox}
\usepackage{makecell}
\usepackage{caption}
\usepackage{epstopdf}
\usepackage{multirow}
\usepackage{multicol}
\usepackage{subcaption}
\usepackage{comment}
\usepackage{pifont}
\usepackage{orcidlink}

\begin{document}

\title{Evidential Domain Adaptation for Remaining Useful Life Prediction with Incomplete Degradation

}



\author{ Yubo Hou\,\orcidlink{0000-0002-0084-4380},
          Mohamed Ragab\,\orcidlink{0000-0002-2138-4395},
          Yucheng Wang\,\orcidlink{0000-0002-8679-3851},
          Min Wu\,\orcidlink{0000-0003-0977-3600},~\IEEEmembership{Senior Member, IEEE, }
          Abdulla Alseiari\,\orcidlink{0009-0001-3302-1719},
          Chee-Keong Kwoh\,\orcidlink{0000-0002-8547-6387},
          Xiaoli Li, \IEEEmembership{Fellow, IEEE, }
          Zhenghua Chen*, \IEEEmembership{Senior Member, IEEE}
 \thanks{Yubo Hou, Yucheng Wang, Min Wu, Xiaoli Li and Zhenghua Chen are with Institute for Infocomm Research (I$^2$R), Agency for Science, Technology and Research (A*STAR), Singapore.}
 \thanks{Yubo Hou, Chee-Keong Kwoh and Xiaoli Li are with School of Computer Science and Engineering, Nanyang Technological University, Singapore.}
 \thanks{Mohamed Ragab and Abdulla Alseiari are with Propulsion and Space Research Center, Technology Innovation Institute, UAE.}
  \thanks{*Corresponding author: Zhenghua Chen}
 }


\def\BibTeX{{\rm B\kern-.05em{\sc i\kern-.025em b}\kern-.08em
    T\kern-.1667em\lower.7ex\hbox{E}\kern-.125emX}}
\markboth{IEEE Transactions on Instrumentation and Measurement,~Vol.~X, No.~X, X~X}
{Author \MakeLowercase{\textit{et al.}}: Title}
\maketitle

\begin{abstract}
Accurate Remaining Useful Life (RUL) prediction without labeled target domain data is a critical challenge, and domain adaptation (DA) has been widely adopted to address it by transferring knowledge from a labeled source domain to an unlabeled target domain. Despite its success, existing DA methods struggle significantly when faced with incomplete degradation trajectories in the target domain, particularly due to the absence of late degradation stages. This missing data introduces a key extrapolation challenge.
When applied to such incomplete RUL prediction tasks, current DA methods encounter two primary limitations. First, most DA approaches primarily focus on global alignment, which can misaligns late degradation stage in the source domain with early degradation stage in the target domain. Second, due to varying operating conditions in RUL prediction, degradation patterns may differ even within the same degradation stage, resulting in different learned features. As a result, even if degradation stages are partially aligned, simple feature matching cannot fully align two domains.
To overcome these limitations, we propose a novel evidential adaptation approach called EviAdapt, which leverages evidential learning to enhance domain adaptation. The method first segments the source and target domain data into distinct degradation stages based on degradation rate, enabling stage-wise alignment that ensures samples from corresponding stages are accurately matched. To address the second limitation, we introduce an evidential uncertainty alignment technique that estimates uncertainty using evidential learning and aligns the uncertainty across matched stages.
The effectiveness of EviAdapt is validated through extensive experiments on the C-MAPSS, N-CMAPSS and PHM2010 datasets. Results show that our approach significantly outperforms state-of-the-art methods, demonstrating its potential for tackling incomplete degradation scenarios in RUL prediction. Our code is available via https://github.com/keyplay/EviAdapt.

\end{abstract}


\IEEEpeerreviewmaketitle




\section{Introduction}
\label{introduction}
Prognostics and health management (PHM) of industrial systems and equipment play a crucial role in enhancing reliability, reducing maintenance costs, and improving safety and operational performance \cite{kordestani2023overview}. Within the PHM domain, Remaining Useful Life (RUL) prediction is a pivotal task for making well-informed maintenance decisions. Currently, various approaches have been proposed for RUL prediction, which can be broadly categorized into three types: model-based approaches, data-driven approaches, and hybrid approaches. Specifically, model-based approaches rely on mathematical or physics-based models to describe the degradation behavior of a system, requiring a strong theoretical understanding \cite{9237975}. However, as mechanical systems become increasingly complex, predicting RUL using model-based methods becomes exceedingly challenging. With the growing availability of data from deployed sensors, data-driven approaches~\cite{9354649, 9606678, 9705208} have gained popularity for RUL prediction.

\begin{figure}[]
\centering
 \includegraphics[width=0.5\textwidth]{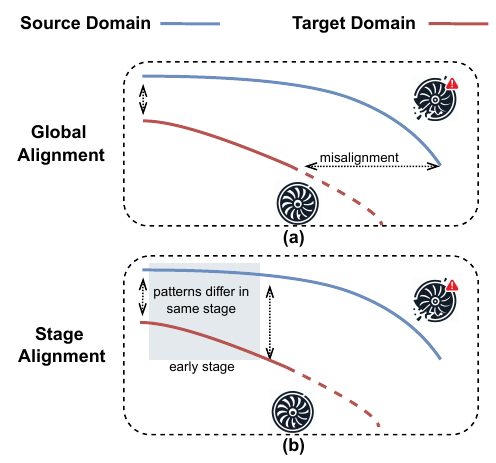}
\caption{Comparison of existing solutions and the proposed solution and the proposed solution for incomplete degradation domain adaptation in RUL prediction. (a) Under global alignment, early degradation in the target domain may be incorrectly aligned with late stage degradation in the source domain. (b) The patterns of the two domains may differ in the early stage, where the source domain experiences steady degradation, while the target domain degrades rapidly.}
\label{fig:problem_solution_intro}
\end{figure}

Despite the promise of data-driven approaches, their success primarily hinges on the assumption of identically distributed data~\cite{9733347}. However, given the dynamics of real-world environments, models are typically trained under one operating condition and tested under another, leading to significant performance degradation due to domain shift. Furthermore, collecting annotated data in new operating conditions and retraining the model is both impractical and costly.
Given these challenges, accurately predicting RULs under various working conditions with limited labeled data poses significant difficulties. To address these obstacles, unsupervised domain adaptation (UDA) has emerged a promising technique that facilitates knowledge transfer from a labeled source domain to a distinct yet related unlabeled target domain~\cite{liu2022deep}. Recently, there has been an increased focus on UDA for RUL prediction task, aiming to learn domain-invariant features by reducing domain shift either through adversarial training~\cite{ragab2020adversarial, ragab2022self, da2020remaining} or minimizing the statistical distance between domains~\cite{9788003, cheng2021transferable, mao2019predicting}.

Although current DA methods have proven effective in addressing domain shift in RUL prediction, they are typically designed under the assumption that complete run-to-failure data is available in the target domain. However, in industrial systems, such data is often scarce due to safety concerns, as most systems are not allowed to operate until failure. As a result, the target domain may lack crucial data from the final degradation stage, leaving only data from the early degradation stages available. When existing DA methods are applied in these incomplete settings, they often struggle for two main reasons. First, by not considering the progression of degradation stages, these methods typically achieve global alignment, leading to misalignment across domains. As illustrated in Fig. \ref{fig:problem_solution_intro} (a), early degradation in the target domain may align with late degradation in the source domain, causing misalignment. Second, due to varying operating conditions, degradation patterns of different machines may differ even within the same degradation stage~\cite{li2021degradation}. As shown in Fig. \ref{fig:problem_solution_intro} (b), the patterns of the two domains differ in the early stage, where the source domain experiences steady degradation, while the target domain degrades rapidly. These differing patterns lead to different feature representations. Strict alignment of such features, even when degradation stages are properly aligned, can negatively impact overall alignment performance.


To address these challenges, we propose a novel evidential adaptation approach, EviAdapt, for RUL prediction with incomplete degradation data in the target domain. To address the misalignment of degradation stage, we introduce a novel approach that segments both source and target domains into distinct degradation stages based on degradation rate, followed by stage-wise alignment of samples. By aligning samples at same degradation stages, our method ensures accurate degradation alignment across domains, addressing a critical limitation in existing DA techniques for RUL prediction.
To resolve the issue of strict feature alignment, we propose an evidential uncertainty alignment technique that focuses on aligning uncertainty levels between corresponding degradation stages rather than directly aligning features. Given that uncertainty is a second-order statistical equivalent~\cite{10.5555/3367471.3367576}, the consistency in uncertainty levels across corresponding degradation stages can effectively bridge differences between domains, thereby improving the alignment in RUL tasks.
Through extensive experiments, we have thoroughly evaluated the performance of our proposed EviAdapt method in accurately predicting the RUL of machines across diverse operating conditions.


The main contributions of this study are listed as follows.
\begin{itemize}
     \item We propose a stage-wise alignment strategy that aligns sample within the same degradation stage across different domains, effectively addressing the misalignment of degardation stage of incomplete lifecycle data in the target domain.

     \item We introduce a novel evidential uncertainty alignment technique that aligns uncertainty levels between corresponding degradation stages.

     \item We conduct extensive experiments on the C-MAPSS, N-CMAPSS and PHM2010 datasets to demonstrate that our EviAdapt approach significantly outperforms existing state-of-the-art methods in cross-domain RUL prediction, validating the effectiveness of our strategies in practical scenarios.

\end{itemize}

\section{Related works}
\label{related_works} 

\subsection{Unsupervised Domain Adaptation for RUL Prediction}
\label{UDA_RUL}
Unsupervised Domain Adaptation (UDA) for RUL prediction aims to mitigate the labeling cost by training neural networks to transfer knowledge from a labeled source domain to an unlabeled target domain. Existing UDA methods strive to achieve high performance on the target domain by minimizing the domain discrepancy. These methods can be categorized into two distinct branches, i.e., metric-based methods and adversarial-based methods.

Metric-based methods enable networks to learn invariant features by enforcing metric constraints. Deep domain confusion (DDC)~\cite{jia2017assessment} employs the maximum mean discrepancy (MMD) to address the challenge of domain discrepancy. Correlation alignment (CORAL)~\cite{sun2017correlation} focused on minimizing the covariance shift between the feature distributions of the source and target domains. 


Adversarial-based methods employ domain discriminator networks to compel the feature extractor to acquire representations that are invariant across domains. Domain adversarial neural network~\cite{da2020remaining} utilized a reverse gradient strategy to conduct adversarial training for both the domain classifier and the feature extractor. Adversarial domain adaption approach for remaining useful life prediction (ADARUL)~\cite{ragab2020adversarial} utilized a conventional GAN loss with flipped labels to learn domain-invariant features. Contrastive adversarial domain adaptation (CADA)~\cite{ragab2020contrastive} incorporated a contrastive loss to persevere target-specific information for RUL prediction. 


The above methods operate under the assumption that the target domain possesses complete run-to-failure data. In reality, acquiring such data in the target domain can be challenging, rendering these methods less effective in practical scenarios.
In response to the condition of incomplete target data, Cons DANN~\cite{siahpour2022novel} utilized a consistency-based regularization term during alignment to mitigate the negative impact of missing information in the incomplete target domain dataset. 
In~\cite{cheng2023remaining}, the authors proposed a generative adversarial network to generate various types of full-cycle degradation data. However, the aforementioned methods treat training data as a homogeneous whole and overlook the discrepancies in characteristics across different degradation stages between the source and target domains, potentially leading to negative transfer effects. 
In~\cite{li2023partial}, the authors proposed an adversarial learning strategy combined with a source-domain instance-weighted degradation fusion scheme for similar degradation levels. However, this method is specifically designed for bearings and lacks generalizability, particularly when applied to aero engine data.
Moreover, all these methods ignore the uncertainty when learning the domain distribution. 

\subsection{Uncertainty Quantification}
Various methods have been developed to estimate uncertainty, including weights reparametrization~\cite{blundell2015weight, kingma2015variational}, dropout~\cite{gal2016dropout}, and ensembling~\cite{lakshminarayanan2017simple}. Although these methods are effective, they are computationally demanding. Evidential learning can estimate uncertainty using single deterministic models. Several works~\cite{sensoy2018evidential, malinin2018predictive, malinin2019reverse} have been proposed for classification tasks. For regression tasks, deep evidential regression~\cite{amini2020deep} was introduced. However, a notable limitation is the reliance on Gaussian assumptions, which may restrict its application. While existing methods primarily rely on evidential learning for uncertainty quantification, our approach leverages uncertainty alignment as a means to reduce the domain gap between source and target.
\section{Methodology}
\label{sec:methodology}

\begin{figure*}[t]
\centering
 \includegraphics[width=1.01\textwidth]{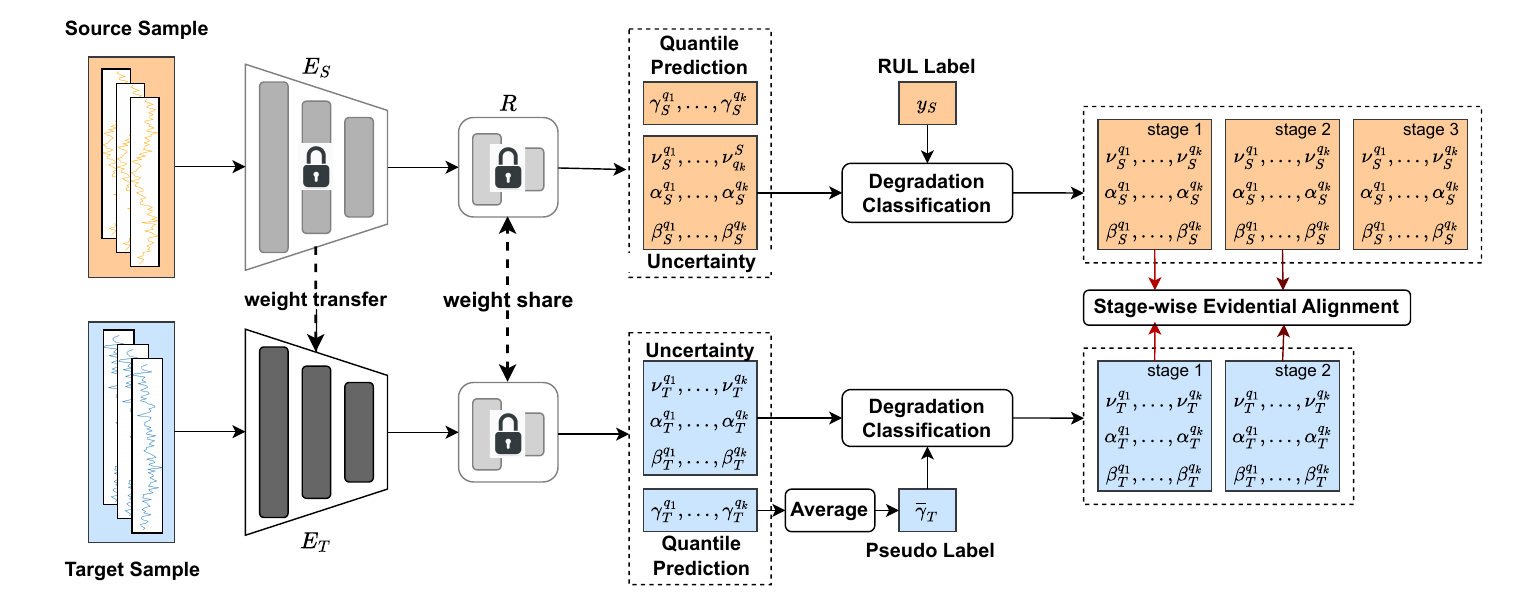}
\caption{An overview of our proposed EviAdapt approach. EviAdapt comprises three main components: source encoder $E_S$, target encoder $E_T$ and shared predictor $R$. $E_S$ and $R$ are pretrained to learn the RUL distribution and its uncertainty of the source domain using evidential learning. During adaptation, source and target data are segmented into different degradation stages. Eventually, $E_T$ is trained to align the uncertainty of the same degradation stages between the source and target domains.}
\label{fig:Unc_method}
\end{figure*}

\subsection{Problem Formulation}
We denote a source domain with $N_S$ labeled samples $\{X^i_S, y^i_S\}^{N_S}_{i=1}$ and a target domain with $N_T$ unlabeled samples $\{X^i_T\}^{N_T}_{i=1}$, where $X^i_S \in \mathbb{R}^{M \times L}$ and $X^i_T \in \mathbb{R}^{M \times L}$ are both multivariate time series data consisting of $M$ sensors and $L$ time steps. $y^i_S$ is the RUL label. We aim to transfer knowledge from labeled source domain to unlabeled target domain and then improve the performance of RUL prediction on the target. Table \ref{tab:notation} provides a summary of the notations employed in this paper.

\begin{table}[]
\centering
\caption{List of notations.}
\label{tab:notation}
\begin{tabular}{ll}
\hline
Notation & Definition                      \\ \hline
 $X_S/X_T$        & source/target data              \\ \hline
 $y_S$        & source RUL label              \\ \hline
 $N_S/N_T$       & number of source/target samples \\ \hline
 $f_S/f_T$        & source/target features          \\ \hline
 $E_S/E_T$        & source/target encoder           \\ \hline
 $R$        & predictor            \\ \hline
 $M$        & number of sensors               \\ \hline
 $L$        & sequence length                 \\ \hline
\end{tabular}
\end{table}

\subsection{Overview}
The overall structure of our proposed EviAdapt method is illustrated in Figure \ref{fig:Unc_method}. EviAdapt comprises three main components: source encoder $E_S$, target encoder $E_T$ and shared predictor $R$. First, we pretrain source encoder $E_S$ and predictor $R$ to learn the RUL distribution and its uncertainty of the source domain using evidential learning. Second we segment source and target data into different degradation stages. Eventually, we train target encoder $E_T$ to align the uncertainty of the same degradation stages between the source and target domains, leveraging the well-trained $E_S$ and $R$ to facilitate the process.

\subsection{Evidential Pretraining on Source Domain}
The first step in our proposed method is to pretrain source encoder and predictor. During this phase, the objective is to train an evidential model using the labeled source domain data to learn the uncertainty of the source domain. Deep evidential regression~\cite{amini2020deep} is a single deterministic forward-pass model that estimates uncertainty under the assumption of Gaussian distribution.
This assumption limits the modeling applications. To overcome this limitation, we employ an evidential Bayesian quantile regression model as the RUL predictor~\cite{huttel2023deep}.
Specifically, a source encoder $E_S$ and a predictor $R$ are trained on source data $X_S$. The source encoder extracts features from source data: $f_S=E_S(X_S)$. The predictor estimates the quantile of the RUL value and its uncertainty based on the extracted features. 


Assuming that $y_S$ come from a Gaussian distribution parameterized in the form of quantile regression, we place a Gaussian prior on the unknown mean and an Inverse-Gamma prior on the unknown variance~\cite{huttel2023deep}:

\begin{equation}
y_S \sim \mathcal{N}(\mu+\tau z, \omega \sigma^2 z), \mu \sim \mathcal{N}(\gamma, \sigma^2 \nu^{-1}), \sigma^2 \sim \Gamma^{-1}(\alpha, \beta)
\end{equation}
where $\Gamma(\cdot)$ is the Gamma function, $\gamma \in \mathbb{R}, \nu>0, \alpha>1, \beta>0$, $\tau = \frac{1-2q}{q(1-q)}$ and $\omega = \frac{2}{q(1-q)}$ are quantile-specific constants from a specified quantile $q$, and $z \sim \mathrm{Exp}\left(\frac{1}{\beta/(\alpha-1)}\right)$.

Together, the distributions of $\mu$ and $\sigma$ form the Normal-Inverse-Gamma (NIG) evidential prior~\cite{amini2020deep}:
\begin{multline}
p(\mu, \sigma^2 | \gamma, \nu, \alpha, \beta) = \\
\frac{\beta^\alpha \sqrt{\nu}}{\Gamma(\alpha)\sqrt{2\pi\sigma^2}}(\frac{1}{\sigma^2})^{\alpha+1} exp\Bigl\{-\frac{2\beta+\nu(\gamma-\mu)^2}{2\sigma^2}\Bigl\}
\end{multline}
The objective is to infer the parameters $(\gamma, \nu, \alpha, \beta)$ of this evidential distribution. By placing the NIG prior on the likelihood parameters, we can derive an analytical solution that produces a Student-t predictive distribution. To learn the parameters of the evidential distribution, we maximize the likelihood of the Student-t distribution. So we minimize the negative log-likelihood loss during training:
\begin{multline}
\mathcal{L}_{NLL} = \frac{1}{2}log(\frac{\pi}{\nu})-\alpha log(4\beta(1+\omega \overline{z} \nu)) + (\alpha+\frac{1}{2}) \\
(log(y-\gamma-\tau \overline{z})^2\nu+4\beta(1+\omega \overline{z} \nu))+log(\frac{\Gamma(\alpha)}{\Gamma(\alpha+\frac{1}{2})}),
\label{eq:nll_loss}
\end{multline}
where $\overline{z}$ is the mean of $z$.

A tilted loss is used as regularization term to penalize evidence of prediction errors:
\begin{equation}
\mathcal{L}_R = max(q(y-\gamma), (q-1)(y-\gamma))\Phi,
\label{eq:tilted_loss}
\end{equation}
where $\Phi=2\nu + \alpha + 1 / \beta$ is model confidence.

Given a set of quantile value, the source encoder and the RUL predictor are optimized with the negative log-likelihood loss and tilted loss:
\begin{equation}
\mathcal{L} = \sum_{q=1} (\mathcal{L}_{NLL}+\lambda \mathcal{L}_{R}),
\label{eq:total_loss}
\end{equation}

We follow the settings described in~\cite{amini2020deep, huttel2023deep} and enforce the constraints on ($\nu, \alpha, \beta$) using a softplus activation function, with an additional +1 added to $\alpha$ to ensure $\alpha > 1$. A linear activation function is used for $\gamma$.

It is worth noting that the RUL predictor estimates the parameters of the NIG distribution for each quantile value $q$, given a set of quantile values [$q_1, q_2, \dots, q_k$]: 
\begin{equation}
(\boldsymbol{\gamma}_S, \boldsymbol{\nu}_S, \boldsymbol{\alpha}_S, \boldsymbol{\beta}_S)=R(f_S). 
\end{equation}
where $\boldsymbol{\gamma}_S=[\gamma^{q_1}_S,\dots,\gamma^{q_k}_S]$, $\boldsymbol{\nu}_S=[\nu^{q_1}_S,\dots,\nu^{q_k}_S]$, $\boldsymbol{\alpha}_S=[\alpha^{q_1}_S,\dots,\alpha^{q_k}_S]$ and $\boldsymbol{\beta}^S=[\beta^{q_1}_S,\dots,\beta^{q_k}_S]$. And RUL value can be estimated by average of $\boldsymbol{\gamma}_S$.

\subsection{Stage Segmentation}

Given the pretrained source model, the focus lies in achieving adaptation upon the unlabeled incomplete target data. Instead of aligning feature, we align uncertainty between source and target domains. 
By aligning the uncertainty, we can effectively align conditional distributions, leading to improved domain alignment.
However, globally aligning uncertainty overlooks incomplete target domain data situation, leading to suboptimal alignment. To address this issue, we propose a stage segmentation to ensure the alignment of corresponding degradation stages during the adaptation phase. Specifically, we classify the degradation of complete source domain and incomplete target domain into different stages based on their degradation speed.


\subsubsection{Identifying Source Degradation Stages}
\begin{figure}[]
\centering
 \includegraphics[width=0.5\textwidth]{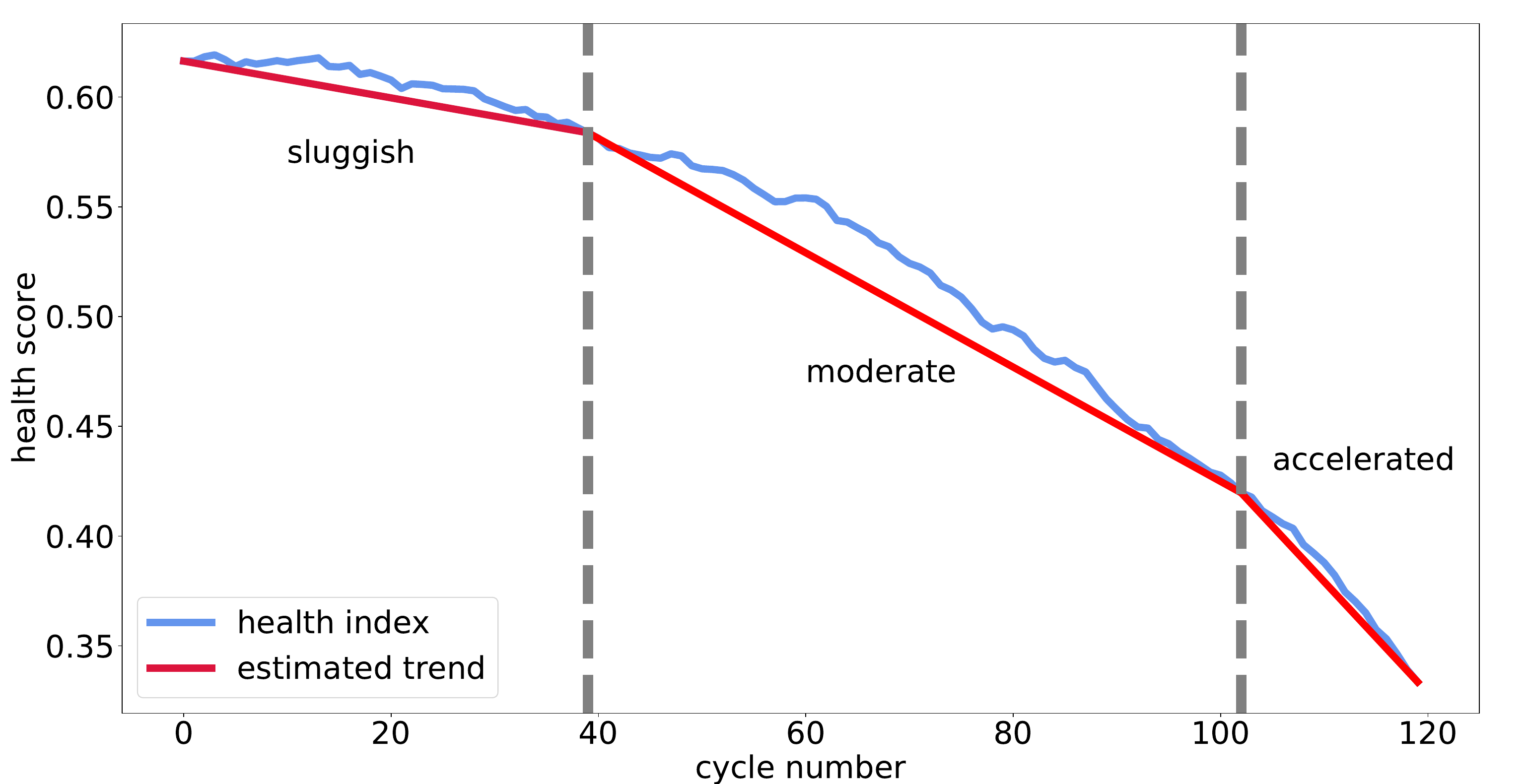}
\caption{Three degradation stages categorized by the health index.}
\label{fig:health_idx}
\end{figure}
For the complete source domain, we categorize the data into three stages using the available RUL labels. To accurately define the boundary for each stage, we calculate an engine's ``health index" (HI) by forming a linear combination of the key sensor readings, following the methods described in \cite{lei2018machinery, liu2021prediction}. 
This curve effectively reflects the overall evolution of the engine’s health from normal operation to failure. After obtaining the HI, we examine how it changes over time for each engine, combined with the RUL information, to observe the varying degradation rates in different sections of the lifecycle.
Based on the health index, we empirically establish life-cycle ranges of (0, 33\%), (33\%, 85\%), and (85\%, 100\%) to denote the sluggish, moderate, and accelerated degradation stages, respectively~\cite{lei2018machinery, liu2021prediction}. The reasons for choosing these three ranges are as follows:
\begin{itemize}
    \item Sluggish stage:
        The HI curve changes relatively slowly here, indicating that the engine is in a more ``healthy'' state with a lower degradation rate.
    \item Moderate stage:
        As operation time increases, the degradation rate begins to accelerate, though it has not yet entered a high-failure-risk phase. The HI value often shows a marked decline compared to the initial stage.
    \item Accelerated stage:
        Near failure, the degradation rate increases significantly, and the HI curve typically exhibits a rapid drop, indicating a steep rise in failure risk.
\end{itemize}

Figure \ref{fig:health_idx} presents the health index of a single engine across its 120 operational cycles. The full cycle is divided into three distinct stages: the sluggish stage (cycles 0 to 40, corresponding to 0–33\%), the moderate stage (cycles 40 to 102, corresponding to 33–85\%), and the accelerated stage (cycles 102 to 120, corresponding to 85–100\%). The figure clearly shows that the trends in the health index vary significantly across these stages, highlighting the differences in degradation behavior.

\subsubsection{Identifying Target Degradation Stages}
For the incomplete target domain, segmenting into the different degradation stages becomes a challenge due to the absence of labels. To address this, we label the target data using the pretrained source model. Due to the incomplete data, it is highly possible that parts of the moderate and fully accelerated degradation stages are missing. Therefore, we classified the data into two stages using pseudo labels. 
We empirically determine life-cycle ranges of (0, 70\%) and (70\%, 100\%) to represent the sluggish and moderate stages, respectively.

\subsection{Stage-wise Evidential Alignment}
After segmentation of the degradation stages, we propose a stage-wise evidential alignment loss $\mathcal{L}_{SEA}$ to align uncertainty levels between corresponding degradation stages rather than directly matching features. The two inputs to this loss are the parameters of the evidential distributions (NIG distributions in our case) from the corresponding degradation stages in the source and target domains.
\begin{equation}
\mathcal{L}_{SEA} = -\sum_{n=1}^{2}\mathbb{E} [k((\boldsymbol{\nu}_S, \boldsymbol{\alpha}_S, \boldsymbol{\beta}_S)_{n}, (\boldsymbol{\nu}_T, \boldsymbol{\alpha}_T, \boldsymbol{\beta}_T)_{n})].
\label{eq:stage_unc_loss}
\end{equation}
$(\boldsymbol{\nu}, \boldsymbol{\alpha}, \boldsymbol{\beta})_{n}$ represents the parameters of the evidential distribution associated with degradation stage $n$, where $n=1$ corresponds to the sluggish stage and $n=2$ to the moderate stage. The $k(\cdot, \cdot)$ is a kernel function to measure the distance between the evidential parameters from the source and the target domain.

\subsection{Overall Objective}
In the EviAdapt algorithm (Algorithm \ref{al:EvA}), the primary objective is to fine-tune the target encoder $E_T$ for the RUL estimation of equipment. As indicated in Line 1, the algorithm begins with the pretraining of the source encoder $E_S$ and the predictor $R$ using the source domain data $(X_S, y_S)$. After pretraining, the target domain data $X_T$ are passed through the pretrained source encoder $E_S$, and the predictor $R$ generates the pseudo labels $\hat{y_T}$ (Line 2 and 3).
Next, the source domain data $X_S$ is segmented into three distinct degradation stages, based on the source labels $y_S$ (Line 4). Similarly, the target domain data $X_T$ is segmented into two degradation stages, based on the pseudo labels $\hat{y_T}$ (Line 5).
Then the parameters of the evidential distribution $\boldsymbol{\nu}_S, \boldsymbol{\alpha}_S, \boldsymbol{\beta}_S$ for the segmented source domain data are estimated by the predictor $R$ according to the input samples (Line 6 and 7).
During the training iteration, the parameters of the evidential distribution $\boldsymbol{\nu}_T, \boldsymbol{\alpha}_T, \boldsymbol{\beta}_T$ for the segmented target domain data are estimated (Line 9 and 10).
The source and target parameters of the evidential distribution serve as inputs to the stage-wise evidential alignment loss 
$\mathcal{L}_{SEA}$, which is designed to align uncertainty levels between corresponding degradation stages from the source and target domains, thereby training the target encoder $E_T$ for better adaptation.
(Line 11 and 12). Finally, the well trained target encoder $E_T$ can be used in predicting RUL in the target domain.

\begin{algorithm}[h]
\DontPrintSemicolon
\KwIn{Source domain: $\{X_S, y_S\}$,
      Target domain: $X_T$}
\KwOut{Trained target encoder $E_T$}
$E_S, R \gets pretrain(X_S, y_S)$\;
$\boldsymbol{\gamma}_T, \boldsymbol{\nu}_T, \boldsymbol{\alpha}_T, \boldsymbol{\beta}_T \gets R(E_S(X_T))$\;
$\hat{y_T} \gets$ average $\boldsymbol{\gamma}_T$ over [$q_1, q_2, \dots, q_k$]\;
$X_{S1}, X_{S2}, X_{S3} \gets$ stage segmentation for $X_S$ based on $y_S$\;
$X_{T1}, X_{T2} \gets$ stage segmentation for $X_T$ based on $\hat{y_T}$\;
$\boldsymbol{\gamma}_{S1}, \boldsymbol{\nu}_{S1}, \boldsymbol{\alpha}_{S1}, \boldsymbol{\beta}_{S1} \gets R(E_S(X_{S1}))$\;
$\boldsymbol{\gamma}_{S2}, \boldsymbol{\nu}_{S2}, \boldsymbol{\alpha}_{S2}, \boldsymbol{\beta}_{S2} \gets R(E_S(X_{S2}))$\;
\While{iteration}{
  $\boldsymbol{\gamma}_{T1}, \boldsymbol{\nu}_{T1}, \boldsymbol{\alpha}_{T1}, \boldsymbol{\beta}_{T1} \gets R(E_T(X_{T1}))$\;
  $\boldsymbol{\gamma}_{T2}, \boldsymbol{\nu}_{T2}, \boldsymbol{\alpha}_{T2}, \boldsymbol{\beta}_{T2} \gets R(E_T(X_{T2}))$\;
  $loss \gets \mathcal{L}_{SEA}((\boldsymbol{\nu}_{S}, \boldsymbol{\alpha}_{S}, \boldsymbol{\beta}_{S})_n, (\boldsymbol{\nu}_{T}, \boldsymbol{\alpha}_{T}, \boldsymbol{\beta}_{T})_n)$\;
  $\theta^{(t+1)}_{E_T} = \theta^{(t)}_{E_T} - lr \cdot \nabla_{\theta_{E_T}}loss$
   //Update $E_T$ by minimizing $loss$ through backpropagation
}
\Return{$E_T$}
\caption{Our Proposed EviAdapt}
\label{al:EvA}
\end{algorithm}

\section{Experiments}
\label{experiments}

\subsection{Data Preparation}
We employ the C-MAPSS, N-CMAPSS and PHM2010 benchmark datasets to evaluate the performance of State-of-the-Art methods. 
\begin{itemize}
    \item \emph{C-MAPSS}: This dataset comprises operational data from four different turbofan engines, each functioning under unique operational conditions and exhibiting specific fault modes, as detailed in Table~\ref{tab:dataset}. It features readings from 21 sensors placed strategically to monitor engine health. To simulate the situation of incomplete target domain data, we excluded the final 40\% of the run-to-failure data for each engine in the target training set but kept the full test data, then applied the preprocessing methodology from \cite{ragab2020contrastive}, resulting in a refined dataset that includes data from 14 selected sensors, with labels indicating the engines' remaining useful life.
    \item \emph{N-CMAPSS}: This dataset~\cite{data6010005} documents the run-to-failure trajectories of turbofan engines. Unlike the C-MAPSS dataset, which is confined to standard cruise phase conditions, N-CMAPSS includes simulations of entire flight cycles—climb, cruise, and descent phases—thereby enhancing the fidelity of degradation modeling. These improvements make N-CMAPSS better equipped to capture the complex dynamics present in real systems. For our experiments, we utilize datasets DS01, DS02, and DS03, which provide data from 20 channels, as outlined in Table~\ref{tab:dataset}. Similarly, we excluded the final 40\% of the run-to-failure data for each engine in the target training set while retaining the full test data. The rest preprocessing follows the methodology described in \cite{mo2022multi}.
    \item \emph{PHM2010}: The PHM2010 dataset~\cite{PHM2010} provides detailed records of cutting tool wear during machining processes. Due to the strong correlation between the RUL of the cutter and wear, and given that the dataset labels represent the wear depth, the objective is to predict the wear depth of the cutters after each cut~\cite{qin2022real}. For our experiments, we leverage datasets C1, C4, and C6, as these records include labels for evaluation. Each record, representing the continuous use of the same cutter, contains approximately 315 cutting instances with 7 sensor channels.
    Additionally, a sliding window has been employed to preprocess the data. As the data is relatively sparse while the dataset is large, a large stride of 1000 has been used to cover the broad range of the whole dataset. Meanwhile, the sliding window of 100 is adopted to reduce computational overhead. 
    The data is normalized to scale all features to the range [0, 1]. For domain adaptation tasks, the first 60\% of the cutting instances from the records are used as the target training set, while the remaining 40\% are reserved for testing.
\end{itemize}

\begin{table*}[]
\centering
\caption{Details of benchmark datasets.}
\label{tab:dataset}
\begin{tabular}{l|cccc|ccc|ccc}
\hline
Dataset & \multicolumn{4}{c|}{C-MAPSS} & \multicolumn{3}{c|}{N-CMAPSS} & \multicolumn{3}{c}{PHM2010}\\ \hline
Sub-dataset      & FD001 & FD002 & FD003 & FD004  & DS01 & DS02 & DS03 & C1 & C4 & C6 \\ \hline
\# Engine units for training        & 100   & 260   & 100   & 249  & 6   & 6   & 9  &  \multicolumn{3}{c}{NA}\\ \hline
\# Engine units for testing       & 100   & 259   & 100   & 248  & 4   & 3   & 6 & \multicolumn{3}{c}{NA} \\ \hline
\# Complete Training samples        & 17731 & 48558 & 21220 & 56815 & 4881 & 5237 & 5532 & 69176 & 70270 & 69054\\ \hline
\# Incomplete Training samples     & 9438 & 24607 & 11893 & 29434  & 2918 & 3131 & 3302  & 42162 & 41505 &41432\\ \hline
\# Testing samples         & 100   & 259   & 100   & 248   & 2717   &  1240  & 4225 & 28108   & 27671  & 27622\\ \hline
\end{tabular}
\end{table*}

\subsection{Experimental Setting}
All experiments run five times and the average results are shown to prevent the effect of random initialization. For fair comparisons, we adopt the same LSTM~\cite{hochreiter1997long} feature extractor for proposed method and baseline methods.
Due to the variation between different dataset, different number of layers and different size of hidden states are selected for each dataset. Table~\ref{tab:para_lstm} shows the detailed feature extractor parameters for each dataset. Additionally, we set batch size as 256, optimizer as Adam, learning rate as 5e-5 for the target encoder. We use quantile values [0.25, 0.75] in our proposed method to compare with the state-of-the-art methods. Furthermore, we built and trained our model based on Pytorch and NVIDIA GeForce RTX A4000 GPU. We adopt root mean square error (RMSE) and Score \cite{ragab2020contrastive} as evaluation metrics for the C-MAPSS and N-CMAPSS datasets. Notably, as the Score function has parameters specifically designed for turbofan engine datasets and are not suitable for the PHM2010 dataset, only RMSE has been utilized for measurement. The lower the two indicators are, the better the model is.

\begin{table}[]
\centering
\caption{Parameter setting for the LSTM feature extractor.}
\label{tab:para_lstm}
\begin{tabular}{l|ccc}
\hline
Parameter       & C-MAPSS & N-CMAPSS &PHM2010 \\ \hline
\# of Layers        & 5   & 1   & 5    \\ \hline
\# of Hidden Size       & 32   & 64   & 32     \\ \hline
Dropout        & 0.5 & 0.1 &0.1 \\ \hline
\end{tabular}
\end{table}

The RMSE metric is defined as follows:
\begin{equation}
RMSE = \sqrt{\frac{1}{N}\sum_{i=1}^N (y_i-\widehat{y_i})^2},
\label{eq:rmse}
\end{equation}
where $\widehat{y_i}$ and $y_i$ represent the estimated RUL and true RUL respectively.

The RMSE metric assigns equal importance to both early and late RUL predictions. However, in prognostics applications, late RUL predictions have more detrimental consequences for the systems. In order to address this concern, the Score metric is employed, which imposes a more severe penalty for late RUL predictions. The Score metric is expressed as follows:
\begin{equation}
Score_i =\left\{\begin{aligned}
 e^{-\frac{\widehat{y_i}-y_i}{13}}-1; \widehat{y_i} < y_i, \\
 e^{\frac{\widehat{y_i}-y_i}{10}}-1; \widehat{y_i} > y_i, 
\end{aligned}\right.
\label{eq:score_i}
\end{equation}

\begin{equation}
Score =\sum_{i=1}^N Score_i.
\label{eq:score}
\end{equation}

\subsection{Comparison with State-of-the-Art Methods}

\begin{table*}[]
\caption{Comparison of the proposed EviAdapt against benchmark approaches on C-MAPSS (RMSE). Note that F1 is short for FD001, and F1$\rightarrow$F2 refers to the scenario where FD001 is the source domain and FD002 is the target domain. Bold indicates the best result, and underline indicates the second-best result.}
\label{tab:cmpass_result_rmse}
\centering
\setlength{\tabcolsep}{5pt}
\begin{NiceTabular}{l|c|c|c|c|c|c|c|c|c|c|c|c|c}
\toprule
Methods & F1$\rightarrow$F2             & F1$\rightarrow$F3             & F1$\rightarrow$F4             & F2$\rightarrow$F1             & F2$\rightarrow$F3             & F2$\rightarrow$F4             & F3$\rightarrow$F1             & F3$\rightarrow$F2             & F3$\rightarrow$F4             & F4$\rightarrow$F1             & F4$\rightarrow$F2             & F4$\rightarrow$F3             & Avg.           \\ \midrule 
Source-RMSE & 55.35 & 60.52 &	55.98 &	52.29	& 53.60	& 54.86&	41.94&	44.37&	40.57 &	46.09&	51.46&	49.10&	50.51 \\
DDC        & 40.00 & 40.06 & 43.98 & 38.83 & 48.24 & 43.06 & 41.55 & 40.65 & 43.68 & 40.51 & 39.61 & 38.16 & 41.53 \\
Deep Coral & 36.90 & 41.75 & 45.21 & 35.88 & 41.16 & 43.85 & 36.16 & 37.08 & 37.78 & 36.11 & 36.80 & 35.96 & 38.72 \\
ADARUL     & 44.73 & 54.37 & 36.93 & 48.41 & 48.84 & 49.38 & 34.19 & 36.19 & 39.98 & 28.05 & 33.76 & 37.43 & 41.02 \\
CADA       & 45.24 & 54.57 & 38.73 & 49.60 & 49.07 & 48.74 & 41.23 & 39.41 & 40.55 & 32.22 & 39.65 & 38.17 & 43.10 \\
Cons DANN  & \textbf{28.52} & \ul{34.65} & \textbf{29.95} & \ul{22.86} & 28.77 & 29.63 & \ul{21.02} & \ul{24.32} & \textbf{27.44} & \textbf{21.21} & \ul{25.40} & \ul{23.67} & \ul{26.45} \\ \midrule 
Source-EVI & 59.00     & 65.83     & 56.42     & 56.45     & 57.77     & 55.30     & 54.84     & 55.48     & 48.53     & 46.14     & 51.34     & 52.43     & 54.96 \\
EviAdapt &\ul{29.95} & \textbf{31.13} & \ul{33.72} & \textbf{21.64} & \textbf{26.03} & \textbf{28.83} & \textbf{19.19} & \textbf{19.92} & \ul{28.53} & \ul{23.67} & \textbf{18.59} & \textbf{20.58} & \textbf{25.00}        \\ \bottomrule     
\end{NiceTabular}
\end{table*}

\begin{table*}[]
\caption{Comparison of the proposed EviAdapt against benchmark approaches on C-MAPSS (Score). Note that F1 is short for FD001, and F1$\rightarrow$F2 refers to the scenario where FD001 is the source domain and FD002 is the target domain. Bold indicates the best result, and underline indicates the second-best result.}
\label{tab:cmpass_result_score}
\centering
\setlength{\tabcolsep}{5pt}
\begin{NiceTabular}{l|c|c|c|c|c|c|c|c|c|c|c|c|c}
\toprule
Methods & F1$\rightarrow$F2             & F1$\rightarrow$F3             & F1$\rightarrow$F4             & F2$\rightarrow$F1             & F2$\rightarrow$F3             & F2$\rightarrow$F4             & F3$\rightarrow$F1             & F3$\rightarrow$F2             & F3$\rightarrow$F4             & F4$\rightarrow$F1             & F4$\rightarrow$F2             & F4$\rightarrow$F3             & Avg.           \\ \midrule 
Source-RMSE  &75683 &	46833&	72564&	16834&	23400&	67866&	6740&	25030&	19995 &	11831&	46169&	13971&	35576 \\
DDC        & 22208 & 6368  & 21318 & 4930  & 6514  & 16958 & 6140  & 22422 & 21397 & 5939  & 23050 & 4238  & 13457 \\
Deep Coral & 11782 & 6723  & 18697 & 2986  & 6802  & 16098 & 3001  & 11538 & 9907  & 3033  & 11495 & 3416  & 8790  \\
ADARUL     & 36618 & 27370 & 19401 & 11467 & 15171 & 44955 & 3862  & 16706 & 19380 & 3207 & 8913  & 6032 & 17757 \\
CADA       & 37815 & 27805 & 24187 & 13201 & 15690 & 47750 & 6676  & 22293 & 19964 & 4907 & 13183 & 6560 & 20003 \\
Cons DANN  & \ul{5817}  & \ul{4614}  & \textbf{7028}  & \ul{1164}  & \ul{2017}  & \ul{5948}  & \textbf{1076}  & \ul{6019}  & \ul{5811}  & \textbf{947}   & \ul{4529}  & \ul{1343}  & \ul{3859} \\ \midrule
Source-EVI & 109644    & 78206     & 91815     & 26023  & 34167  & 93708  & 20639     & 76765     & 42292     & 13700     & 47593     & 20632     & 54599 \\
EviAdapt & \textbf{4976}  & \textbf{2435}  & \ul{7456}  & \textbf{908}   & \textbf{1404}  & \textbf{5227}  & \ul{1178}  & \textbf{3974}  & \textbf{5146}  & \ul{1361}  & \textbf{3980}  & \textbf{809}   & \textbf{3238}   \\ \bottomrule     
\end{NiceTabular}
\end{table*}

\begin{table*}[]
\caption{Comparison of the proposed EviAdapt against benchmark approaches on N-CMAPSS (RMSE). Bold indicates the best result, and underline indicates the second-best result.}
\label{tab:ncmpass_result_rmse}
\centering
\setlength{\tabcolsep}{5pt}
\begin{NiceTabular}{l|c|c|c|c|c|c|c}
\toprule
Methods         & DS01$\rightarrow$DS02 & DS01$\rightarrow$DS03 & DS02$\rightarrow$DS01 & DS02$\rightarrow$DS03 & DS03$\rightarrow$DS01 & DS03$\rightarrow$DS02 & Avg. \\ \midrule 
Source-RMSE & 21.89 & 28.23 & 33.29 & 25.64 & 40.37 & 26.50 & 29.32  \\
DDC        & 19.23 & 21.12 & 21.05 & 15.86 & 24.75 & 17.78 & 19.96 \\
Deep Coral & 14.06 & 15.75 & 21.34 & 16.37 & 24.42 & 16.21 & 18.03 \\
ADARUL     & \ul{9.97}  & 13.50 & 17.51 & 14.20 & \ul{14.07} & \ul{9.18}  & \ul{13.07} \\
CADA       & 10.05 & 14.12 & \ul{16.07} & \ul{13.14} & 18.88 & 12.90 & 14.19 \\
Cons DANN  & 10.03 & \ul{11.91} & 16.52 & 12.96 & 17.48 & 10.25 & 13.19 \\ \midrule
Source-EVI & 23.64 & 27.89 & 30.68 & 24.25 & 40.23 & 28.11  & 29.14 \\
EviAdapt & \textbf{9.23}  & \textbf{9.87}  & \textbf{9.69}  & \textbf{9.74}  & \textbf{9.87}  & \textbf{5.81}  & \textbf{9.04} \\ \bottomrule 
\end{NiceTabular}
\end{table*}

\begin{table*}[]
\caption{Comparison of the proposed EviAdapt against benchmark approaches on N-CMAPSS (Score). Bold indicates the best result, and underline indicates the second-best result.}
\label{tab:ncmpass_result_score}
\centering
\setlength{\tabcolsep}{5pt}
\begin{NiceTabular}{l|c|c|c|c|c|c|c}
\toprule
Methods         & DS01$\rightarrow$DS02 & DS01$\rightarrow$DS03 & DS02$\rightarrow$DS01 & DS02$\rightarrow$DS03 & DS03$\rightarrow$DS01 & DS03$\rightarrow$DS02 & Avg. \\ \midrule 
Source-RMSE & 7202 & 58571 & 62615 & 49246 & 118257 & 11480 & 51228 \\
DDC        & 5767  & 28992 & 16178 & 14566 & 24616 & 5325  & 15908 \\
Deep Coral & 2880 & 13444 & 17037 & 15810 & 22697 & 3804 & 12613 \\
ADARUL     & \ul{1579} & 11370 & 11924 & 11135 & \ul{5908} & \ul{1441} & 7226 \\
CADA       & 1632 & 12260 & \ul{8568} & 9814 & 13335 & 2459 & 8011 \\
Cons DANN  & 1715 & \ul{8216} & 9634 & \ul{9675} & 10183 & 1783 & \ul{6868}  \\  \midrule
Source-EVI & 8662 & 55812 & 53767 & 45930 & 119890 & 14051 & 49685 \\
EviAdapt & \textbf{1515} & \textbf{5742} & \textbf{3804} & \textbf{7553} & \textbf{4205} & \textbf{979} & \textbf{3966}         \\ \bottomrule 
\end{NiceTabular}
\end{table*}

\begin{table*}[]
\caption{Comparison of the proposed EviAdapt against benchmark approaches on PHM2010 (RMSE). Bold indicates the best result, and underline indicates the second-best result.}
\label{tab:phm2010_result_rmse}
\centering
\setlength{\tabcolsep}{5pt}
\begin{NiceTabular}{l|c|c|c|c|c|c|c}
\toprule
Methods & C1$\rightarrow$C4  &  C1$\rightarrow$C6& C4$\rightarrow$C1 & C4$\rightarrow$C6 & C6$\rightarrow$C1    &  C6$\rightarrow$C4  & Avg.           \\ \midrule 
Source-RMSE  & 0.209 & 0.111 & 0.311 & 0.201 & 0.280 & 0.312 & 0.237 \\
DDC        & 0.176 & 0.272 & 0.223 & 0.340 & 0.127 & \ul{0.170} & 0.218\\
Deep Coral & \ul{0.175} & 0.194 & 0.155 & 0.150 & \ul{0.116} & \textbf{0.169} & 0.160  \\
ADARUL     & 0.216 & 0.114 & 0.313 & 0.202 & 0.300 & 0.323 & 0.245 \\
CADA       & 0.209 & 0.111 & 0.311 & 0.201 & 0.280 & 0.312 & 0.237 \\
Cons DANN  & \textbf{0.173} & \ul{0.106} & \ul{0.109} & \ul{0.118} & \textbf{0.112} & 0.233 & \ul{0.142}  \\  \midrule 
Source-EVI  & 0.219 & 0.112 & 0.200 & 0.129 & 0.259 & 0.300 & 0.203 \\
EviAdapt & 0.185 & \textbf{0.100} & \textbf{0.105} & \textbf{0.108} & 0.145 & 0.192 & \textbf{0.139}\\ \bottomrule     
\end{NiceTabular}
\end{table*}

We compare the proposed EviAdapt with a range of state-of-the-art UDA methods, including conventional domain adaptation methods like DDC~\cite{tzeng2014deep}, Coral~\cite{sun2017correlation}, ADARUL~\cite{ragab2020adversarial}, and CADA~\cite{ragab2020contrastive}, as well as Cons DANN~\cite{siahpour2022novel} which is specifically designed for incomplete domain adaptation in the context of RUL prediction. Further, the results of source only (Source) are also compared. Due to the use of evidential learning for pre-training in the proposed method, the last layer of our predictor differs from other domain adaptation methods. Therefore, we present two sets of source-only results. The first set, Source-RMSE, uses a predictor trained with the RMSE loss. The second set, Source-EVI, employs an evidential predictor trained with negative log-likelihood loss and tilted loss.

Table~\ref{tab:cmpass_result_rmse} shows the RMSE results and Table~\ref{tab:cmpass_result_score} shows the Score results respectively in 12 cross-domain scenarios for RUL prediction on C-MAPSS dataset. From the results, we observe that even though Source-EVI performs significantly worse than Source-RMSE, EviAdapt achieves the best performance in the 8 scenarios for RMSE and in the 9 scenarios for Score. Notably, EviAdapt improves the average performance by over 5\% in RMSE and 16\% in Score compared to the second-best method.

Similarly, Table~\ref{tab:ncmpass_result_rmse} shows the RMSE and Table~\ref{tab:ncmpass_result_score} shows the Score results respectively in 6 cross-domain scenarios for RUL prediction on N-CMAPSS dataset. From the results, we observe that EviAdapt achieves the best performance across all scenarios with regards to both RMSE and Score. Notably, EviAdapt improves the average performance by over 30\% in RMSE and 42\% in Score compared to the second-best method. 

Moreover, Table~\ref{tab:phm2010_result_rmse} shows the RMSE results in 6 cross-domain scenarios for wear depth prediction on PHM2010 dataset. From the results, we observe that EviAdapt achieves the best performance in the 3 scenarios. Notably, EviAdapt improves the average performance by over 2\% in RMSE compared to the second-best method.

These consistent superior performances demonstrate EviAdapt's ability to align uncertainty of the same degradation stage effectively, leading to significant advancements in RUL prediction across different domains.

\subsection{Ablation Study}

To validate the contribution of key components, we conducted an ablation study on our proposed EviAdapt using the C-MAPSS dataset. We derived two variants of EviAdapt based on different combinations of alignment scope and alignment type. The alignment scope refers to the range over which the alignment is applied and includes two kinds: global alignment (“$G$”), which considers the entire dataset, and same degradation stage alignment (“$S$”), which focuses on aligning data within the same degradation stage. The alignment type refers to the specific aspect of the data being aligned and includes two kinds: alignment by feature (“$Fea$”) and alignment by uncertainty (“$Unc$”). It is worth noting that the combination (“$S$”, “$Unc$”) corresponds to our proposed method.

Table~\ref{tab:ablation_rmse} and Table~\ref{tab:ablation_score} presents the comparative outcomes between EviAdapt and its variants. Our observations reveal that alignment by uncertainty surpasses alignment by feature, showing an improvement of 65\% in terms of average Score. 
Furthermore, same degradation stage alignment outperforms global alignment, with a maximum improvement of 67\% in terms of average Score. These results underscore the effectiveness of stage-wise alignment and alignment by uncertainty.

\begin{table*}[]
\caption{Ablation study for the proposed EviAdapt on C-MAPSS (RMSE).}
\label{tab:ablation_rmse}
\centering
\setlength{\tabcolsep}{3pt}
\begin{NiceTabular}{c|c|c|c|c|c|c|c|c|c|c|c|c|c|c}
\toprule
\begin{tabular}[c]{@{}c@{}}Alignment\\ Scope\end{tabular} & \begin{tabular}[c]{@{}c@{}}Alignment\\ Type\end{tabular}    & F1$\rightarrow$F2             & F1$\rightarrow$F3             & F1$\rightarrow$F4             & F2$\rightarrow$F1             & F2$\rightarrow$F3             & F2$\rightarrow$F4             & F3$\rightarrow$F1             & F3$\rightarrow$F2             & F3$\rightarrow$F4             & F4$\rightarrow$F1             & F4$\rightarrow$F2             & F4$\rightarrow$F3             & Avg.           \\ \midrule
$G$ & $Fea$   &  \textbf{25.47}     & \textbf{27.47}     & \ul{34.36}     & 26.11     & 34.98     & 36.39     & 26.87     & 32.80     & 33.67     & 25.42     & 26.74     & 25.00     & 29.61 \\
$G$ & $Unc$   & \ul{29.61}     & 32.91     & 36.00     & \ul{23.81}     & \ul{31.53}     & \ul{31.50}     & \ul{24.35}     & \ul{24.72}     & \textbf{26.45}     & \textbf{23.08}     & \ul{24.12}     & \ul{23.83}     & \ul{27.66}   \\ 
$S$ & $Unc$   &29.95 & \ul{31.13} & \textbf{33.72} & \textbf{21.64} & \textbf{26.03} & \textbf{28.83} & \textbf{19.19} & \textbf{19.92} & \ul{28.53} & \ul{23.67} & \textbf{18.59} & \textbf{20.58} & \textbf{25.00}     \\ \bottomrule   
\end{NiceTabular}
\end{table*}

\begin{table*}[]
\caption{Ablation study for the proposed EviAdapt on C-MAPSS (Score).}
\label{tab:ablation_score}
\centering
\setlength{\tabcolsep}{3pt}
\begin{NiceTabular}{c|c|c|c|c|c|c|c|c|c|c|c|c|c|c}
\toprule
\begin{tabular}[c]{@{}c@{}}Alignment\\ Scope\end{tabular} & \begin{tabular}[c]{@{}c@{}}Alignment\\ Type\end{tabular}    & F1$\rightarrow$F2             & F1$\rightarrow$F3             & F1$\rightarrow$F4             & F2$\rightarrow$F1             & F2$\rightarrow$F3             & F2$\rightarrow$F4             & F3$\rightarrow$F1             & F3$\rightarrow$F2             & F3$\rightarrow$F4             & F4$\rightarrow$F1             & F4$\rightarrow$F2             & F4$\rightarrow$F3             & Avg.           \\ \midrule
$G$ & $Fea$   &\textbf{3953}     & \ul{2344}      & 35593     & 7522      & 11296     & 59736     & 27492     & 94580     & 80814     & 3577      & 8686      & 2317      & 28159  \\
$G$ & $Unc$    & 6224      & \textbf{2246}      & \ul{8268}      & \ul{3832}      & \ul{7880}      & \ul{29588}     & \ul{6345}      & \ul{26558}     & \ul{15633}     & \textbf{891}       & \textbf{2608}      & \ul{1093}      & \ul{9264}  \\
$S$ & $Unc$  & \ul{4976}  & 2435  & \textbf{7456}  & \textbf{908}   & \textbf{1404}  & \textbf{5227}  & \textbf{1178}  & \textbf{3974}  & \textbf{5146}  & \ul{1361}  & \ul{3980}  & \textbf{809}   & \textbf{3238}         \\ \bottomrule   
\end{NiceTabular}
\end{table*}

\subsection{Sensitivity Analysis}
\begin{figure}[]
\centering
 \includegraphics[width=0.5\textwidth]{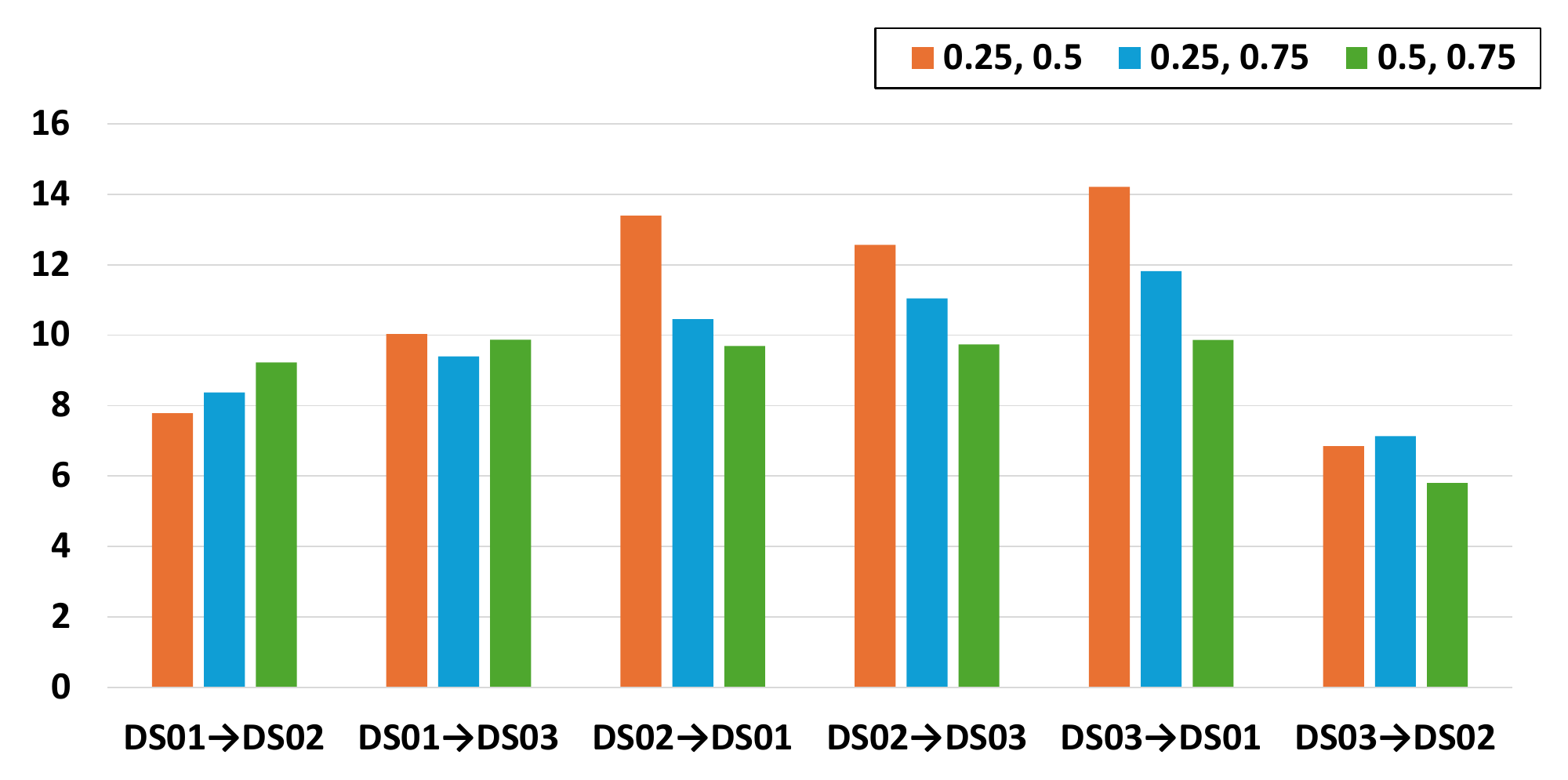}
\caption{The sensitivity analysis for different set of quantile values on N-CMAPSS (RMSE).}
\label{fig:sen_rmse}
\end{figure}

\begin{figure}[]
\centering
 \includegraphics[width=0.5\textwidth]{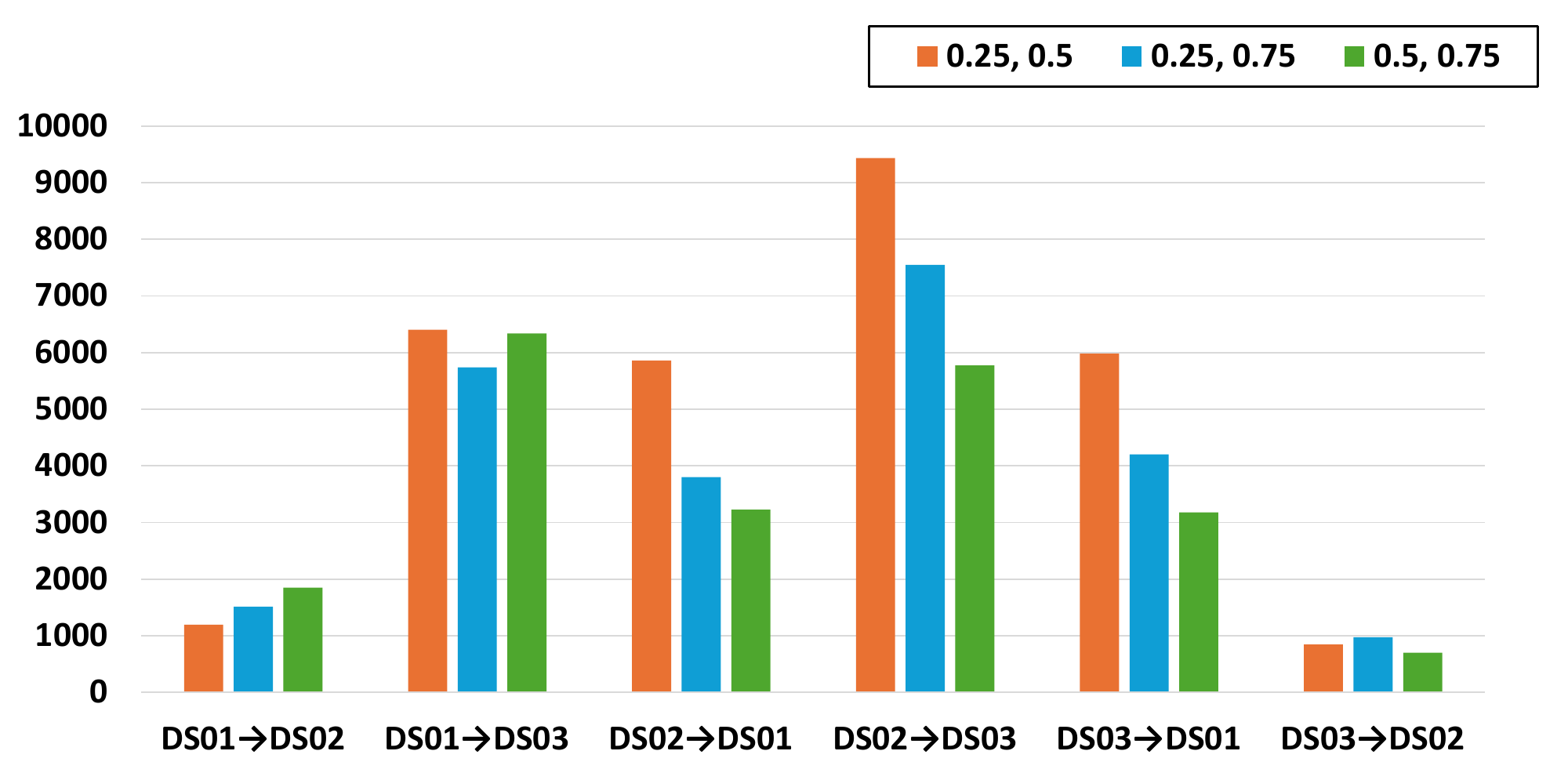}
\caption{The sensitivity analysis for different set of quantile values on N-CMAPSS (Score).}
\label{fig:sen_score}
\end{figure}

We conducted a sensitivity analysis for different set of quantile values $q$ on N-CMAPSS dataset to investigate the impact of evidential learning on the proposed method. Several experiments were carried out using various set of values, including [0.25, 0.5], [0.5, 0.75] and [0.25, 0.75]. Figure~\ref{fig:sen_rmse} and Figure~\ref{fig:sen_score} illustrate that the proposed method demonstrates that the optimal set of quantile values varies across different domains. Specifically, the quantiles [0.25, 0.5] yields the best performance for DS01$\rightarrow$DS02 in terms of RMSE and Score, while [0.25, 0.5] achieves the best performance for DS02$\rightarrow$DS01 and DS02$\rightarrow$DS03 in terms of Score. The impact of the quantile values on domain adaptation varies depending on the source and target domains. The potential reason could be attributed to different sets of quantiles capture varying domain information.

\subsection{Visualization of Feature Distribution}
\begin{figure}[]
    \centering
    \includegraphics[width=0.35\textwidth]{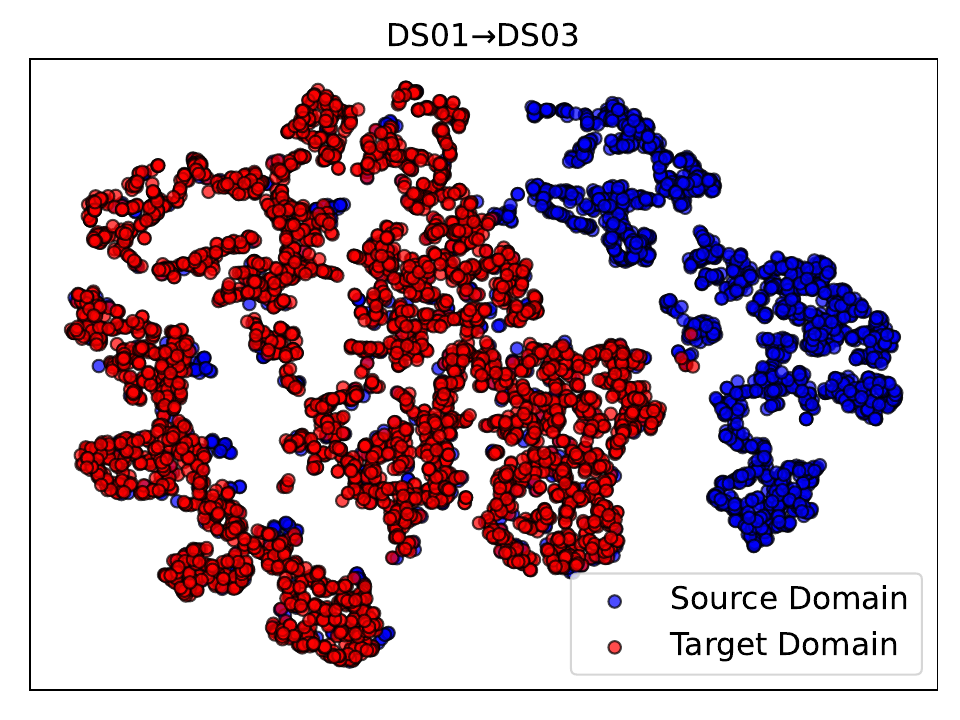}   
    \includegraphics[width=0.35\textwidth]{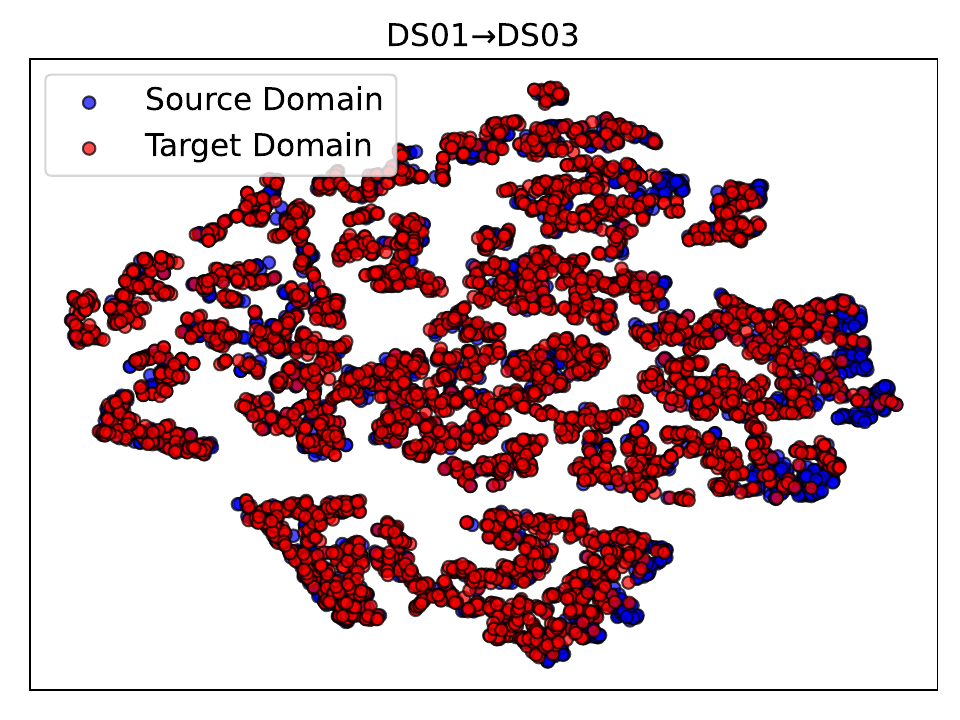} 
    \caption{Feature distribution analysis. Up: before adaptation. Down: after adaptation.}
    \label{fig:tsne}
\end{figure}

To showcase the effectiveness of the proposed method, we employed t-SNE to visualize the latent features in the reduced-dimensional space both before and after adaptation on the N-CMAPSS dataset for the DS01$\rightarrow$DS03 scenario. As illustrated in Fig.\ref{fig:tsne}, prior to adaptation, a considerable portion of the source samples are positioned far from the target distribution, emphasizing the domain gap. In contrast, post-adaptation visualization demonstrates that the source and target distributions are closely aligned. Collectively, these visualizations clearly indicate that our approach successfully minimizes the disparity between the source and target feature distributions.

\section{Conclusion}
\label{conclusion}
In this paper, we found that most existing domain adaptation methods fail under incomplete target domains. To address this, we propose a novel approach called EviAdapt for unsupervised domain adaptation in RUL prediction tasks. Unlike previous methods that overlook the misalignment of degradation stage and inherent uncertainties in RUL tasks, EviAdapt aligns the uncertainty within the same degradation stage by the proposed stage-wise evidential alignment technique, thereby highlighting the limitations of existing methods. 
Through extensive experiments, our results demonstrate the remarkable performance of EviAdapt, surpassing state-of-the-art methods on the C-MAPSS, N-CMAPSS, and PHM2010 datasets, with average improvements of 16\%, 42\%, and 2\%, respectively.

In future work, we plan to develop a source-free domain adaptation (SFDA) method for RUL prediction, addressing the challenge of reducing dependencies on fully labeled source domains. In many industrial settings, production data is highly sensitive and often contains confidential information, making it infeasible to share raw data across domains due to corporate privacy policies. Instead, SFDA focuses on adapting pretrained models to target domains without requiring direct access to the source data, ensuring compliance with privacy constraints while maintaining effective domain adaptation.
 


\bibliographystyle{IEEEtran}
\bibliography{references}

@article{liu2021prediction,
  title={Prediction of remaining useful life of multi-stage aero-engine based on clustering and LSTM fusion},
  author={Liu, Junqiang and Lei, Fan and Pan, Chunlu and Hu, Dongbin and Zuo, Hongfu},
  journal={Reliability Engineering \& System Safety},
  volume={214},
  pages={107807},
  year={2021},
  publisher={Elsevier}
}

@article{huttel2023deep,
  title={Deep Evidential Learning for Bayesian Quantile Regression},
  author={H{\"u}ttel, Frederik Boe and Rodrigues, Filipe and Pereira, Francisco C{\^a}mara},
  journal={arXiv preprint arXiv:2308.10650},
  year={2023}
}

@inproceedings{blundell2015weight,
  title={Weight uncertainty in neural network},
  author={Blundell, Charles and Cornebise, Julien and Kavukcuoglu, Koray and Wierstra, Daan},
  booktitle={International conference on machine learning},
  pages={1613--1622},
  year={2015},
  organization={PMLR}
}

@article{li2023partial,
  title={Partial domain adaptation in remaining useful life prediction with incomplete target data},
  author={Li, Xiang and Zhang, Wei and Li, Xu and Hao, Hongshen},
  journal={IEEE/ASME Transactions on Mechatronics},
  year={2023},
  publisher={IEEE}
}

@article{lakshminarayanan2017simple,
  title={Simple and scalable predictive uncertainty estimation using deep ensembles},
  author={Lakshminarayanan, Balaji and Pritzel, Alexander and Blundell, Charles},
  journal={Advances in neural information processing systems},
  volume={30},
  year={2017}
}

@article{sensoy2018evidential,
  title={Evidential deep learning to quantify classification uncertainty},
  author={Sensoy, Murat and Kaplan, Lance and Kandemir, Melih},
  journal={Advances in neural information processing systems},
  volume={31},
  year={2018}
}

@article{malinin2018predictive,
  title={Predictive uncertainty estimation via prior networks},
  author={Malinin, Andrey and Gales, Mark},
  journal={Advances in neural information processing systems},
  volume={31},
  year={2018}
}

@article{malinin2019reverse,
  title={Reverse kl-divergence training of prior networks: Improved uncertainty and adversarial robustness},
  author={Malinin, Andrey and Gales, Mark},
  journal={Advances in Neural Information Processing Systems},
  volume={32},
  year={2019}
}

@article{liu2022deep,
  title={Deep unsupervised domain adaptation: A review of recent advances and perspectives},
  author={Liu, Xiaofeng and Yoo, Chaehwa and Xing, Fangxu and Oh, Hyejin and El Fakhri, Georges and Kang, Je-Won and Woo, Jonghye and others},
  journal={APSIPA Transactions on Signal and Information Processing},
  volume={11},
  number={1},
  year={2022},
  publisher={Now Publishers, Inc.}
}

@inproceedings{gal2016dropout,
  title={Dropout as a bayesian approximation: Representing model uncertainty in deep learning},
  author={Gal, Yarin and Ghahramani, Zoubin},
  booktitle={international conference on machine learning},
  pages={1050--1059},
  year={2016},
  organization={PMLR}
}

@article{kingma2015variational,
  title={Variational dropout and the local reparameterization trick},
  author={Kingma, Durk P and Salimans, Tim and Welling, Max},
  journal={Advances in neural information processing systems},
  volume={28},
  year={2015}
}

@article{amini2020deep,
  title={Deep evidential regression},
  author={Amini, Alexander and Schwarting, Wilko and Soleimany, Ava and Rus, Daniela},
  journal={Advances in neural information processing systems},
  volume={33},
  pages={14927--14937},
  year={2020}
}

@ARTICLE{9606678,
  author={Chen, Chuang and Lu, Ningyun and Jiang, Bin and Xing, Yin and Zhu, Zheng Hong},
  journal={IEEE Transactions on Instrumentation and Measurement}, 
  title={Prediction Interval Estimation of Aeroengine Remaining Useful Life Based on Bidirectional Long Short-Term Memory Network}, 
  year={2021},
  volume={70},
  number={},
  pages={1-13}}

@ARTICLE{9237975,
  author={Hou, Bingchang and Wang, Dong and Wang, Yi and Yan, Tongtong and Peng, Zhike and Tsui, Kwok-Leung},
  journal={IEEE Transactions on Instrumentation and Measurement}, 
  title={Adaptive Weighted Signal Preprocessing Technique for Machine Health Monitoring}, 
  year={2021},
  volume={70},
  number={},
  pages={1-11}}

@ARTICLE{9788003,
  author={Ma, Guijun and Xu, Songpei and Yang, Tao and Du, Zhenbang and Zhu, Limin and Ding, Han and Yuan, Ye},
  journal={IEEE Transactions on Neural Networks and Learning Systems}, 
  title={A Transfer Learning-Based Method for Personalized State of Health Estimation of Lithium-Ion Batteries}, 
  year={2024},
  volume={35},
  number={1},
  pages={759-769}
}

@article{ragab2022self,
  title={Self-supervised autoregressive domain adaptation for time series data},
  author={Ragab, Mohamed and Eldele, Emadeldeen and Chen, Zhenghua and Wu, Min and Kwoh, Chee-Keong and Li, Xiaoli},
  journal={IEEE Transactions on Neural Networks and Learning Systems},
  volume={35},
  number={1},
  pages={1341--1351},
  year={2022},
  publisher={IEEE}
}

@article{li2021degradation,
  title={Degradation alignment in remaining useful life prediction using deep cycle-consistent learning},
  author={Li, Xiang and Zhang, Wei and Ma, Hui and Luo, Zhong and Li, Xu},
  journal={IEEE Transactions on Neural Networks and Learning Systems},
  year={2021},
  publisher={IEEE}
}

@Article{data6010005,
AUTHOR = {Arias Chao, Manuel and Kulkarni, Chetan and Goebel, Kai and Fink, Olga},
TITLE = {Aircraft Engine Run-to-Failure Dataset under Real Flight Conditions for Prognostics and Diagnostics},
JOURNAL = {Data},
VOLUME = {6},
YEAR = {2021},
NUMBER = {1},
ARTICLE-NUMBER = {5},
ISSN = {2306-5729},
DOI = {10.3390/data6010005}
}

@article{qin2022real,
  title={Real-time remaining useful life prediction of cutting tools using sparse augmented Lagrangian analysis and Gaussian process regression},
  author={Qin, Xiao and Huang, Weizhi and Wang, Xuefei and Tang, Zezhi and Liu, Zepeng},
  journal={Sensors},
  volume={23},
  number={1},
  pages={413},
  year={2022},
  publisher={MDPI}
}

@misc{PHM2010,
  title = {PHM Society 2010 PHM Society Conference Data Challenge},
  year = {2010},
  url = {https://phmsociety.org/phm_competition/2010-phm-society-conference-data-challenge/}
}

@article{hochreiter1997long,
  title={Long short-term memory},
  author={Hochreiter, Sepp and Schmidhuber, J{\"u}rgen},
  journal={Neural computation},
  volume={9},
  number={8},
  pages={1735--1780},
  year={1997},
  publisher={MIT press}
}

@inproceedings{10.5555/3367471.3367576,
author = {Wen, Jun and Zheng, Nenggan and Yuan, Junsong and Gong, Zhefeng and Chen, Changyou},
title = {Bayesian uncertainty matching for unsupervised domain adaptation},
year = {2019},
isbn = {9780999241141},
publisher = {AAAI Press},
booktitle = {Proceedings of the 28th International Joint Conference on Artificial Intelligence},
pages = {3849–3855},
numpages = {7},
location = {Macao, China},
series = {IJCAI'19}
}

@article{cheng2023remaining,
  title={Remaining useful life prediction combined dynamic model with transfer learning under insufficient degradation data},
  author={Cheng, Han and Kong, Xianguang and Wang, Qibin and Ma, Hongbo and Yang, Shengkang and Xu, Kun},
  journal={Reliability Engineering \& System Safety},
  volume={236},
  pages={109292},
  year={2023},
  publisher={Elsevier}
}

@article{siahpour2022novel,
  title={A novel transfer learning approach in remaining useful life prediction for incomplete dataset},
  author={Siahpour, Shahin and Li, Xiang and Lee, Jay},
  journal={IEEE Transactions on Instrumentation and Measurement},
  volume={71},
  pages={1--11},
  year={2022},
  publisher={IEEE}
}

@inproceedings{mo2022multi,
  title={Multi-objective optimization of extreme learning machine for remaining useful life prediction},
  author={Mo, Hyunho and Iacca, Giovanni},
  booktitle={Applications of Evolutionary Computation: 25th European Conference, EvoApplications 2022, Held as Part of EvoStar 2022, Madrid, Spain, April 20--22, 2022, Proceedings},
  pages={191--206},
  year={2022},
  organization={Springer}
}

@article{ragab2020contrastive,
  title={Contrastive adversarial domain adaptation for machine remaining useful life prediction},
  author={Ragab, Mohamed and Chen, Zhenghua and Wu, Min and Foo, Chuan Sheng and Kwoh, Chee Keong and Yan, Ruqiang and Li, Xiaoli},
  journal={IEEE Transactions on Industrial Informatics},
  volume={17},
  number={8},
  pages={5239--5249},
  year={2020},
  publisher={IEEE}
}

@article{tzeng2014deep,
  title={Deep domain confusion: Maximizing for domain invariance},
  author={Tzeng, Eric and Hoffman, Judy and Zhang, Ning and Saenko, Kate and Darrell, Trevor},
  journal={arXiv preprint arXiv:1412.3474},
  year={2014}
}

@article{sun2017correlation,
  title={Correlation alignment for unsupervised domain adaptation},
  author={Sun, Baochen and Feng, Jiashi and Saenko, Kate},
  journal={Domain adaptation in computer vision applications},
  pages={153--171},
  year={2017},
  publisher={Springer}
}

@article{da2020remaining,
  title={Remaining useful lifetime prediction via deep domain adaptation},
  author={da Costa, Paulo Roberto de Oliveira and Ak{\c{c}}ay, Alp and Zhang, Yingqian and Kaymak, Uzay},
  journal={Reliability Engineering \& System Safety},
  volume={195},
  pages={106682},
  year={2020},
  publisher={Elsevier}
}

@ARTICLE{9733347,
  author={Mao, Wentao and Liu, Jing and Chen, Jiaxian and Liang, Xihui},
  journal={IEEE Transactions on Instrumentation and Measurement}, 
  title={An Interpretable Deep Transfer Learning-Based Remaining Useful Life Prediction Approach for Bearings With Selective Degradation Knowledge Fusion}, 
  year={2022},
  volume={71},
  number={},
  pages={1-16}}

@ARTICLE{9705208,
  author={Ren, Likun and Qin, Haiqin and Xie, Zhenbo and Li, Bianjiang and Xu, Kejun},
  journal={IEEE Transactions on Instrumentation and Measurement}, 
  title={Aero-Engine Remaining Useful Life Estimation Based on Multi-Head Networks}, 
  year={2022},
  volume={71},
  number={},
  pages={1-10}}

@ARTICLE{9354649,
  author={Zhao, Huimin and Liu, Haodong and Jin, Yang and Dang, Xiangjun and Deng, Wu},
  journal={IEEE Transactions on Instrumentation and Measurement}, 
  title={Feature Extraction for Data-Driven Remaining Useful Life Prediction of Rolling Bearings}, 
  year={2021},
  volume={70},
  number={},
  pages={1-10}}

@article{lei2018machinery,
  title={Machinery health prognostics: A systematic review from data acquisition to RUL prediction},
  author={Lei, Yaguo and Li, Naipeng and Guo, Liang and Li, Ningbo and Yan, Tao and Lin, Jing},
  journal={Mechanical systems and signal processing},
  volume={104},
  pages={799--834},
  year={2018},
  publisher={Elsevier}
}

@article{mao2019predicting,
  title={Predicting remaining useful life of rolling bearings based on deep feature representation and transfer learning},
  author={Mao, Wentao and He, Jianliang and Zuo, Ming J},
  journal={IEEE Transactions on Instrumentation and Measurement},
  volume={69},
  number={4},
  pages={1594--1608},
  year={2019},
  publisher={IEEE}
}

@inproceedings{ragab2020adversarial,
  title={Adversarial transfer learning for machine remaining useful life prediction},
  author={Ragab, Mohamed and Chen, Zhenghua and Wu, Min and Kwoh, Chee Keong and Li, Xiaoli},
  booktitle={2020 IEEE international conference on prognostics and health management (ICPHM)},
  pages={1--7},
  year={2020},
  organization={IEEE}
}

@article{cheng2021transferable,
  title={Transferable convolutional neural network based remaining useful life prediction of bearing under multiple failure behaviors},
  author={Cheng, Han and Kong, Xianguang and Chen, Gaige and Wang, Qibin and Wang, Rongbo},
  journal={Measurement},
  volume={168},
  pages={108286},
  year={2021},
  publisher={Elsevier}
}

@article{kordestani2023overview,
  title={An Overview of the State-of-the-Art in Aircraft Prognostic and Health Management Strategies},
  author={Kordestani, Mojtaba and Orchard, Marcos E and Khorasan, Khashayar and Saif, Mehrdad},
  journal={IEEE Transactions on Instrumentation and Measurement},
  year={2023},
  publisher={IEEE}
}

@article{jia2017assessment,
  title={Assessment of data suitability for machine prognosis using maximum mean discrepancy},
  author={Jia, Xiaodong and Zhao, Ming and Di, Yuan and Yang, Qibo and Lee, Jay},
  journal={IEEE transactions on industrial electronics},
  volume={65},
  number={7},
  pages={5872--5881},
  year={2017},
  publisher={IEEE}
}

\end{document}